\documentclass[letterpaper, 10 pt, conference]{ieeeconf}  

\IEEEoverridecommandlockouts                              

\usepackage{graphics} 
\usepackage{epsfig} 
\usepackage{mathptmx} 
\usepackage{times} 
\usepackage{amsmath} 
\usepackage{amssymb}  

\newcommand{\R}{{\mathbb R}}

\newcommand{\HQ}{{\mathbb H}}
\newcommand{\HD}{{\mathbb {H}_D}}

  \newtheorem{theo}{Theorem}[section]

  \newtheorem{defn}[theo]{Definition}

  \newtheorem{rem}[theo]{Remark}

\newcommand{\dq}{\,\mathbf{dq}}
\newcommand{\p}{\,\mathbf p}
\newcommand{\q}{\,\mathbf q}

\title{\LARGE \bf
Object Handover Prediction using Gaussian Processes clustered with Trajectory Classification 
}

\author{Muriel Lang, Satoshi Endo, Oliver Dunkley and Sandra Hirche
\thanks{All authors are with the Institute for Information-Oriented Control, Faculty of Electrical Engineering and Information Technology, Technische Universit\"at M\"unchen, D-80290 Munich
{\tt\small muriel.lang@tum.de}}%
}

\begin{document}

\maketitle
\thispagestyle{empty}
\pagestyle{empty}

\begin{abstract}
A robotic system which approximates the user intention and appropriate complimentary motion is critical for successful human-robot interaction. 
Here, we demonstrate robustness of the Gaussian Process (GP) clustered with a stochastic classification technique for trajectory prediction using an object handover scenario. By parametrising real 6D hand movements during human-human object handover using dual quaternions, variations of handover configurations were classified in real-time and then the remaining hand trajectory was predicted using the GP. 
The results highlights that our method can classify the handover configuration at an average of~$43.4\%$ of the trajectory and the final hand configuration can be predicted within the normal variation of human movement. In conclusion, we demonstrate that GPs combined with a stochastic classification technique is a robust tool for proactively estimating human motions for human-robot interaction. 
 
\end{abstract}

\section{INTRODUCTION}

Recognition of action intention and goal of a human user is a powerful tool for designing a proactive robotic assistant in human-robot interaction (HRI). While wearable and ambulatory devices have proven useful for tracking the physical movements of the users in various applications, estimating socio-cognitive processes of the human user remains a significant challenge. In a typical motor coordination between two humans, the dyad's actions are evolved through mutual expectations about each other's task contribution and constraints. During an object handover, for example, people grasp and orient an object so that the receiver can later grasp it optimally with regard to his/her intended use~\cite{Gonzalez2011}. On the other hand, the receiver automatically identifies an appropriate action complimentary to the passer's movement~\cite{Newman-Norlund2007}. In order for a cognitive system to recognise the action intention of the humans, a robust modelling technique for non-linear motions becomes essential. 

Recently, Gaussian processes (GPs) have proven suitable for modelling human movement~\cite{Wang2008}. We extended GPs over dual quaternions and demonstrated robustness of this technique for learning and making predictions about strongly non-linear motions with appreciable uncertainty~\cite{Lang2014}. The adaptation of GPs to perform over dual quaternions allows modelling of 6D motions consisting of large rotational and translational movements. Furthermore, GPs do not only provide a best estimate for the prediction, but also a measure of its uncertainty. Thus, the GP over dual quaternions is a highly potent regression model for learning and predicting non-linear 6D motions such as variable human motion trajectories. 
A downfall of the GP is that the model needs to know reference data of the human user, and it cannot be generalised for context-dependent variations of human movements. While classification methods are often used for categorising trajectories, the state of the art methods, which typically are based on Hidden Markov Models (HMMs)~\cite{Nascimento2010, Kulic2007} can only provide a delayed class nomination. In order to overcome these downfalls, we present how GPs can be hierarchically clustered with a stochastic classification based on the Mahalanobis distance, to proactively identify the action intention of the human and make a trajectory prediction in real-time. 

We briefly explain the GP in section~\ref{GP} and introduce an extension to capture 6D rigid motions parametrised by dual quaternions. In section~\ref{classification}, we present the classification method using a new similarity measure between the observed 6D poses and trajectories expressed with a modified Mahalanobis distance. Finally, in section~\ref{experiment}, the model prediction and classification accuracies are evaluated using real kinematic data in human-human object handover.

\section{GAUSSIAN PROCESS FOR RIGID MOTIONS}\label{GP}

The Gaussian process (GP) is defined in this section, before GP regression over significant rotations is introduced. Then, the GP is extended to 6D rigid motions.

\subsection{Definition of a Gaussian Process}

A GP is a set of random variables, such that every finite subset is jointly Gaussian distributed~\cite{Rasmussen2004}. It is fully specified by mean~$m(\mathbf x)$ and kernel functions~$k(\mathbf x,\mathbf x^\prime)$. The kernel function defines the elements of the covariance matrix~$\mathbf{K}$ by~$(\mathbf{K})_{ij} = k(\mathbf x_i,\mathbf x_j)$. Thus, the GP is defined by
\begin{equation}
	f(x) = \text{GP}(m(\mathbf x),k(\mathbf x,\mathbf x^\prime)).
\end{equation}
Part of the most widely used kernels in machine learning~\cite{Rasmussen2006} is the squared exponential kernel function
\begin{equation}
	k(\mathbf x,\mathbf x^\prime) = \sigma_f^2\ \exp\left( - \frac{d^2(\mathbf x,\mathbf x^\prime)}{2 l^2} \right),
	\label{squaredexponential}
\end{equation}
where~$d(\mathbf x,\mathbf x^\prime) = \|\mathbf x-\mathbf x^\prime\|$ is the Euclidean distance,~$l$ is the length-scale of the hyperparameters and~$\sigma_f$ is the signal noise.

\subsection{Gaussian Process on Pure Rotations}\label{orientations}

For simplicity, this section explains how GPs over pure rotations can be parametrized by unit quaternions. Here, the GP maps unit quaternions representing rotations and corresponding rotational velocities. The GP input is points on the unit sphere~$S_3$, and the output is vectors~${\mathbf v}_{TS}$ in the tangent space~$TS_{\q}$ of the corresponding rotation quaternion~\mbox{$\q= q_w + q_xi + q_yj + q_zk \in \mathbb{H}$}, with parameter entries~$q_w,\,q_x,\,q_y,\,q_z \in \R$ such that~$\|\q\| = 1$. 
\begin{rem}
The unit quaternions are a double coverage of the 3D orientations, i.e. opposite points~$\pm \q$ on the sphere~$S_3$ represent the same orientation. As GP models over quaternions~$\q$ provides the same rotation as GP over~$-\q$, we do not distinguish between the opposite unit quaternions.
\end{rem}
To obtain the rotated quaternion~\mbox{$\q_\text{next}$}, the velocity vector~$\mathbf{v}_{TS}$ is projected to the sphere~$S_3$.
We do this by the central projection
\begin{eqnarray}
&\Pi_{\q}:TS_{\q} \to S_3 \nonumber\\ 
&\Pi_{\q}({\mathbf v}_{TS}) = \{ -\frac{\mathbf{v}}{\| \mathbf{v} \|} , \frac{\mathbf{v}}{\| \mathbf{v} \|} \},
\label{projection}
\end{eqnarray}
where $\mathbf{v} = \q+\mathbf{B}{\mathbf v}_{TS}$. 
Here, the basis~$\mathbf{B}$ is the canonical representation of the 3D tangent space~$TS_{\q}$ in the space~$\R^4$~\cite{Feiten2011}.
Choosing this representation avoids cumbersome learning restrictions in the GP, to ensure that the output is a unit quaternion. Furthermore, it allows defining the Gaussian uncertainty of the velocity vector prediction~${\mathbf v}_{TS}$ in a Euclidean tangent space so that no Gaussian needs defined on the hypersphere~$S_3$. 

The GP on the sphere~$S_3$ is fully specified by a mean function~$m(\mathbf x)$ and a covariance function~$k(\mathbf x,\mathbf x^\prime)$ defined on the unit quaternions. The mean function is restricted to the unit hypersphere~$m:S_3 \to S_3$. 
The kernel function~\mbox{$k:S_3\times S_3\to \R_0^+=\{x\in \R: x\geq 0\}$} is based on the arc length, as quaternion rotations provide motions on the shortest curve on~$S_3$.
The length of the arc section between two quaternions~$\q$ and~$\q^\prime$ is obtained by
\begin{equation}
 d_\text{arc}(\q,\q^\prime) = \min\,\arccos(\langle \pm\q,\pm \q^\prime\rangle),
 \label{darc}
\end{equation}
where~$\langle \q,\q^\prime\rangle = \q\overline{\q^\prime}$ is the scalar product over quaternions.
\begin{rem}
In the metric~$d_\text{arc}$, the length of the shorter great circle section is selected, as the unit quaternions can be restricted to lay on the same hemisphere without loss of generality. 
\end{rem}
The kernel function~$k:S_3\times S_3\to \R_0^+$ over unit quaternions~$\q,\,\q^\prime \in S_3$, is defined by
\begin{equation}
 k_\text{arc}(\q,\q^\prime) = \sigma_f^2\ \exp\left( - \frac{d^2_\text{arc}(\q,\q^\prime)}{2 l ^2} \right),
\label{arc}
\end{equation}
where~$d_\text{arc}$ as in~(\ref{darc}) and the hyperparameters~$\sigma_f,\, l>0$.

Using the Cartesian set product, we can combine the unit sphere~$S_3$ with any Euclidean space~$\R^n$. Thus, GP regression over spaces including rotation and translation becomes straight forward using this method. However, a case specific manual weighting between the elements of different subspaces is required and the Cartesian set product neglects all correlations between the subspaces. Therefore, we propose dual quaternions for representing 6D rigid motions.

\subsection{Gaussian Process on 6D Rigid Motions}\label{Poses}

In this section, we introduce the GP over 6D rigid motions consisting of rotation and translation, parametrized by dual quaternions
\begin{equation}
\mathbb{H}_D = \{\dq\ |\ \dq=\q_\text{re} + \epsilon \q_\text{du} \ \&\  \q_\text{re}, \q_\text{du} \in \mathbb{H}\},
\end{equation}
where~$\epsilon$ is a dual unit which holds~\mbox{$\epsilon^2 = 0$}~\cite{Ata2008}. 

A rigid motion can be represented by a dual quaternion as a combination of a rotation quaternion~$\q_r$ and a translation quaternion~$\q_t$. The quaternion~\mbox{$\q_r \in S_3$} has unit length~\mbox{$\|\q_r\| = 1$} and represents a~3D orientation, whereas the translation vector~\mbox{$\p= (p_x, p_y, p_z)^\top \in \R^3$} is represented by an imaginary quaternion~\mbox{$\q_t = p_x i + p_y j + p_z k\in\HQ$}. With these quaternions the rigid motion~$\dq$ is expressed as
\begin{equation}
 \dq := \q_r + \frac{\epsilon}{2} \q_t\q_r.
\label{DQ_create}
\end{equation}
Detailed dual quaternion calculation rules are shown in~\cite{Feiten2013}.

As of unit quaternions, the GP over dual quaternions is trained on a dataset consisting of input poses in~$\HD$ and its time derivatives as an output vector~$(\mathbf v_{TS},\, \dot\p)^\top$.
The subsequent dual quaternion~$\dq_\text{next}$ is obtained from the velocity vector by applying~(\ref{DQ_create}) to the rotation~\mbox{$\q_r = \Pi_{\q}({\mathbf v}_{TS})$} and the translation~\mbox{$\q_t = \q_{\p} + \q_{\dot\p}$}. 

We set the mean function~\mbox{$m:S_3\times\R^3\to S_3\times\R^3$} to zero, i.e.~$m \equiv (1,0,0,0)^\top + \epsilon (0,0,0,0)^\top$. This is a common technique to save computational effort, without limiting the expressiveness of the GP regression. For the kernel function, we define a distance measure based on the transformation~$\vec{\dq}$, which is applied to the pose~$\dq$, to arrive in~$\dq^\prime$. The transforming dual quaternion is obtained by
\begin{equation}
	\vec{\dq} = \overline{\dq}*\dq^\prime,
\end{equation}
where~$*$ denotes the dual quaternion multiplication.
From this dual quaternion~$\vec{\dq} = \vec{\q_r} +\epsilon\vec{\q_d}$, we retrieve rotation and translation by dual quaternion decomposition using~(\ref{DQ_create}). 
The transformation {mag\-ni\-tude} measure is defined as
\begin{equation}
	d_\text{mag}(\dq,\dq^\prime) = d_\text{arc}(\q_0,\vec{\q}_\text{rot}) + \| \vec{\q}_\text{trans}\|,
	\label{dmag}
\end{equation}
where~$\q_0 = (1,0,0,0)^\top$ denotes the quaternion encoding zero rotation.
Then, we define the squared exponential kernel function~\mbox{$k:(S_3\times\R^3)\times(S_3\times\R^3)\to \R_0^+$} over dual quaternions~$\dq,\,\dq^\prime \in S_3\times\R^3$, as
\begin{equation}
	k_\text{mag}(\dq,\dq^\prime) = \sigma_f^2\ \exp\left( - \frac{d^2_\text{mag}(\dq,\dq^\prime)}{2 l^2} \right),
	\label{kmag}
\end{equation}
where~$d_\text{mag}$ as in~(\ref{dmag}) and the hyperparameters~$\sigma_f,\, l >0$.

\section{TRAJECTORY CLASSIFICATION}\label{classification}

In this section, we describe a classification technique for human-human handover trajectories. This method focuses on predicting class labels of 6D object trajectories~\cite{Lee2008}. Following the state of the art classification method based on maximum likelihood estimation~\cite{Nascimento2010}, we use a class-conditional likelihood term~$p(\xi|\hat{\theta},\,A_n)$ for assigning a new observed motion trajectory~$\xi$ into the set of known conditions~\mbox{$\mathbf{A} = \{A_1,\ldots,A_N\}$}. Here,~$A_n\in\mathbf{A}$ and~\mbox{$\hat{\theta}$} are a set of low level model parameter estimates, which are the optimized hyperparameters,~\mbox{$\hat{\theta} = (\hat\sigma_f,\hat l)$} in case of GP motion prediction inside the conditions.

\subsection{Similarity Measure}

In order to account for natural variabilities of the human movements in forms and durations over repetitions, we introduce a normalised inverted Mahalanobis distance of dual quaternions for measuring the similarity between a new observed trajectory~$\xi_{K+1}$ and a set of training trajectories,~$\{\xi_1, \ldots, \xi_K\}$. 
In this method, the scale of the time step~$\tau$ is kept intact for each trajectory. 
Thus, a trajectory~$\xi^n_j$ is defined over repetitions~$j\in\{1,\ldots,K\}$ and conditions~\mbox{$A_n \in \mathbf A$}. The dual quaternion~$\dq(\tau)$ represents the pose of the object~$\xi^n_j(\tau)$ at each time step~$\tau \in \{ 1,\ldots,T_{j,n}\}$. For a new trajectory~$\xi_{K+1}$, the similarity against the training trajectories is then measured in terms of the pose~$\dq(\tau)$ across all conditions~$n = 1,\ldots,N$.  

\begin{defn}
The Mahalanobis distance of a set of dual quaternions~$\dq_1,\ldots,\dq_K$ from a dual quaternion~\mbox{$\dq\in\HD$} is defined as
\begin{equation}
d_\text{M}(\dq) = \hspace{-.5ex}\sqrt{
\begin{pmatrix}  d_{\text{mag},1} & \hspace{-1.5ex} \dots & \hspace{-1.5ex} d_{\text{mag},K} \end{pmatrix}\hspace{-.5ex}
\begin{pmatrix} 
k_{1,\,1} &\hspace{-1ex}\dots  & \hspace{-1ex}k_{1,\,K}\\
\vdots & \hspace{-1.5ex}\ddots & \hspace{-1ex}\vdots \\ 
k_{K,\,1} &\hspace{-1ex}\dots & \hspace{-1ex} k_{K,\,K} 
\end{pmatrix}^{\hspace{-1.3ex}{-1}}\hspace{-1.3ex}
\begin{pmatrix}  d_{\text{mag},1} \\  \vdots \\  d_{\text{mag},K} \end{pmatrix}
}\hspace{-.5ex},
\end{equation}
where~$d_{\text{mag},j} = d_\text{mag}(\dq,\dq_j)$ as presented in~(\ref{dmag}), the covariance~$k_{i,j} = k_\text{mag}(\dq_i,\dq_j)$ as in~(\ref{kmag}) and~$i,j = 1,\ldots,K$.
\end{defn}

At each trajectory $\xi^n_j$ in condition $A_n$, we select a pose~$\dq_j(\tau^\prime)$ of an arbitrary time stamp~$\tau^\prime$, which is closest to~$\dq(\tau)$ according to~(\ref{dmag}). The inverse Mahalanobis distance,~$d^{-1}_\text{M}(\xi_{K+1}(\tau), n)$, is calculated to measure the similarity between~$\dq(\tau)$ and the set of closes poses of condition~$n$.
To obtain the probability~$p(\xi_{K+1}(\tau)|\hat{\theta},\,A_n)$ for each condition~$A_n$, the sum of the similarity measures across conditions is normalized to 1,
\begin{equation}
p(\xi_{K+1}(\tau)|\hat{\theta},\,A_n) = \frac{d^{-1}_\text{M}(\xi_{K+1}(\tau), n)}{\sum_{i =1}^N d^{-1}_\text{M}(\xi_{K+1}(\tau),i)}.
\label{probattau}
\end{equation}
Then, we obtain the class-conditional likelihood term for a new observed trajectory~$\xi_{K+1}$.
\begin{equation}
p(\xi_{K+1}|\hat{\theta},\,A_n) = \int_0^{T_{K+1,n}} p(\xi_{K+1}(\tau)|\hat{\theta},\,A_n) d\tau,
\label{probtotal}
\end{equation}

\subsection{Classification Rules}\label{decision}
We define four decision rules to classify a new trajectory into a set of conditions; two for nominating and two for eliminating candidates. On the one hand, the classification probability defined in~(\ref{probattau}) is contrasted against pre-set absolute thresholds to nominate or eliminate the candidate conditions when the probability falls outside of these thresholds. On the other hand, we implement a moving window technique in parallel to either nominate or eliminate a condition out of the set of conditions based on the magnitude of their relative probabilities. In this latter method, the integral of the probability within a window of the last $m$ time steps is calculated using~(\ref{probtotal}). The window is then moved along the trajectory in real-time. Therefore, this method initiates after the first $m$ time steps of the trajectory onset. 
The classification is terminated when a condition is nominated. When the elimination criteria are met, procedure resumes without the eliminated condition.

\section{EXPERIMENT}\label{experiment}

\begin{figure}[thpb]
	\centering
	\parbox{3.0in}{\includegraphics[width = 3in]{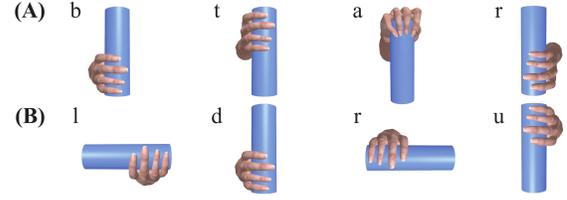}}
	\caption{A list of grasping (A) and handover (B) configurations performed by a passing participant. The grasping movements include \underline{b}ottom, \underline{t}op, (from) \underline{a}bove and \underline{r}eversed, and the handover orientations include \underline{r}ight, \underline{d}own, \underline{l}eft and \underline{u}p.}
 	\label{satoshi}
\end{figure}

In this section we describe an experiment in which a pair of human participants performed variations of object handover, that differ in orientation and position at which a passer grasped the object using a power grip and placed in the receiver's hand. Each condition consisted of a combination of grasping and handover configurations (Figure~\ref{satoshi}), and a total of 10 combinations were used for the experiment. The remaining configurations were omitted as they were biomechanically difficult to perform.

The passer’s upper body motions were tracked using a magnetic motion tracking system (Polhemus Liberty) at~240~Hz, and the motion trajectory, consisting of positions and orientations of the hand were modelled using GPs. For each combination of the grasping and handover configurations, the pair naturally handed over a cylinder (21~cm in length, 5.5~cm in diameter and weighed 280~g) in a blocked design with~20 repetitions. We then trained a GP over dual quaternions using~15 trajectories in each condition, having lengths of~404 to~468 time steps. The hyperparameter learning was performed offline on a commercially available computer with Intel core i5-2500 processor and 16 GB RAM, taking approximately~45 minutes per dimension and condition. 
For classification and motion prediction, we used the 50 remaining trajectories that the GP was not trained on. The procedure was performed on-line in simulation, using the decision rules defined in~\mbox{\ref{decision}}.
When the condition is classified, the GP predicted the remaining trajectory and the handover configuration. 

\begin{figure}[thpb]
	\centering
	\parbox{3.4in}{\includegraphics[width = 3.4in]{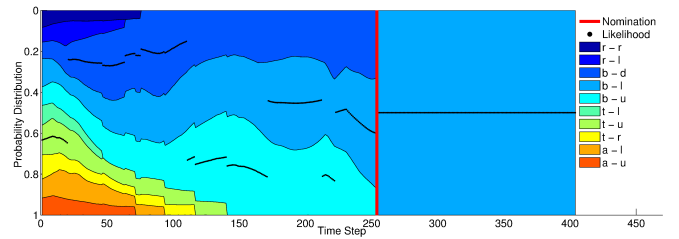}}
	\caption{An example of the classification probability for a \mbox{b-l} trajectory. The red line and the black dots show the nomination step and the maximum likelihood for this condition at the current time step, respectively.}
 	\label{likelihood}
\end{figure}

\begin{figure}[thpb]
	\centering
	\parbox{3.0in}{\includegraphics[width = 3.0in]{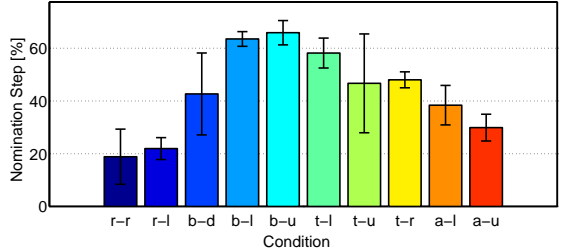}}
	\caption{The average nomination step in percent of the trajectory length is shown for each condition, as well as the covariance of the decision over the 5 test trajectories. The error bars represent one standard deviation of the 5 trails per condition.}
 	\label{bargraph}
\end{figure}

Figure~\ref{likelihood} illustrates an example of classification probability for a b-l trajectory. Here, five conditions are removed when their probabilities reach below the absolute threshold in a time step of~$47,\,64,\,72,\,76$ and~94. Two further conditions are eliminated due to passing the low relative probabilities over windowed integral. At time step 254 ($63\%$ of the trajectory length), the correct condition is classified. As in this example, movements involving a bottom grasp require in general higher decision time-steps than the average due to their particularly similar movements. I total, we achieved~$100\%$ correct classification rate for the 50 testing trajectories on average after~$43.4\%$ of the trajectory.
The average nomination step for each condition is shown in Fig.~\ref{bargraph}.

\begin{figure}[thpb]
	\centering
	\parbox{3.0in}{\includegraphics[width = 3.0in]{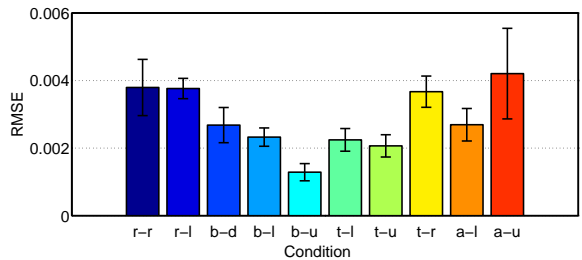}}
	\caption{The average RMSE and the standard deviation from it, using~$d_\text{mag}$ as distance measure is shown for each condition. The error bars represent one standard deviation.}
 	\label{rmse}
\end{figure}

When the classification is complete, we predicted the 6D hand motion using the corresponding GP over dual quaternions and compared our velocity vector predictions to the ground truth trajectories. The rooted mean squared errors~(RMSE) of the dual quaternion velocities are calculated using the distance measure~$d_\text{mag}$ defined in~(\ref{dmag}). In Fig.~\ref{rmse}, the average RMSEs of the GP prediction are visualized for each condition. 

Using the hierarchically clustered stochastic classification method and motion prediction with GPs over dual quaternions, we are able to determine the human action intention and goal at an early stage with high accuracy. 

\section{CONCLUSION AND FUTURE WORK}

We present a technique for modelling human-human object handover using GPs over dual quaternions. Our model is clustered with a stochastic classification technique to recognise the action configuration of the passing participant. The strength of the presented framework is that it accounts for both rotation and position prediction in a single model. The evaluation of the results indicate that the proposed stochastic classification allows for precise and early condition nomination compared to state of the art HMM classification (after~$43.4\%$ of the trajectory in average), followed by accurate motion prediction using GPs over dual quaternions (the overall RMSE is 0.0029 using the~$d_\text{mag}$ distance). 


Our next step is to contrast the speed and reliability of our model prediction against human participants. In the planned experiment, the human would perform a speeded classification task from motion observation and contrast with the results of our model performance. Highlighting the strengths and weakness of our model against human performance would be a valuable information source not only to recognise the intention of the human user but also to model the predictive state of the human partner about robot motion for accomplishing bi-directional intention communication in human-robot interaction.

\addtolength{\textheight}{-12cm}   

\section*{ACKNOWLEDGMENT}

The research leading to these results has received funding from the ERC Starting Grant ``Control based on Human Models (con-humo)" under grant agreement n$^\circ$~337654 and from the European Union Seventh Framework Programme FP7/2007-2013 under grant agreement n$^\circ$~601165 of the project ``WEARHAP - WEARable HAPtics for humans and robots''.

\bibliographystyle{ieeetr}
{\small
\bibliography{bib}
}

\end{document}